\documentclass[letterpaper, 10 pt, conference]{ieeeconf}  % Comment this line out if you need a4paper

\IEEEoverridecommandlockouts
\usepackage{cite}
\usepackage{graphicx}
\usepackage{amsmath,amssymb,amsfonts}
\usepackage{algorithmic}
\usepackage{graphicx}
\usepackage{textcomp}
\usepackage{xcolor}
\usepackage{import}
\usepackage{hyperref}
\usepackage{amsfonts}
\usepackage{comment}
\usepackage{booktabs}
\usepackage{array}
\usepackage{multirow}
\usepackage{multicol}
\usepackage{booktabs}
\usepackage{rotating}
\usepackage[table]{xcolor}
\usepackage[normalem]{ulem}

\def\BibTeX{{\rm B\kern-.05em{\sc i\kern-.025em b}\kern-.08em
    T\kern-.1667em\lower.7ex\hbox{E}\kern-.125emX}}
\begin{document}

\title{\LARGE \bf Not Losing Your Way: Mitigating Visual-Inertial Odometry Drift through On-Demand Visual Feature Creation\thanks{$^*$Corresponding author's email: {\texttt{simone.dangelo@unina.it}}}%
%\thanks{The research leading to these results has been supported by \textcolor{red}{XXX} Project.}% <-this % stops a space
%\thanks{$^*$ Author 1 and Author 2 are both first authors.}%
\thanks{$^{1}$PRISMA Lab, Department of Engineering and Information Technology, University of Naples Federico \uppercase\expandafter{\romannumeral2}, Naples, Italy.}
\thanks{$^{2}$KRUSH Labs B.V., Eindhoven, Netherlands.}
}

\author{%Anonymous Authors
Simone D'Angelo$^{1*}$, Gianmarco Corrado$^{1}$, Federico Esposito$^{1}$, Vincenzo Scognamiglio$^{2}$ and Fabio Ruggiero$^{1}$
}

\maketitle

\begin{abstract}
Visual-inertial odometry (VIO) can degrade during aerial inspection in repetitive or poorly textured environments. We propose an active environment-modification framework for unmanned aerial vehicles (UAVs) relying on cameras and an inertial measurement unit only, without prior maps or additional localization sensors. Inspired by the fairy tale of Tom Thumb, the UAV creates visual features on demand by releasing ground markers or marking walls through physical contact. An information-based indicator triggers interventions using temporal confirmation and hysteresis. Evaluation includes three simulated environments and a $900$-run factorial campaign. Lower cruise speed and active marking are associated with reduced divergence, while accuracy among non-divergent runs depends on speed--strategy interactions. Corridor experiments, with five runs per configuration, use OptiTrack for flight control while VIO-based marking decisions remain active online. The combined framework reduces the reported position root mean square error by approximately $35\%$ relative to baseline VIO, supporting active feature creation for drift mitigation.
\end{abstract}

\begin{comment}
\begin{IEEEkeywords}
Visual-Inertial Odometry, Feature Degradation, Autonomous Exploration, State Estimation, Drift Reduction.
\end{IEEEkeywords}
\end{comment}

\section{Introduction}\label{sec1}
Aerial inspection in global navigation satellite system (GNSS)-denied environments can require unmanned aerial vehicles (UAVs) to traverse tunnels and corridors with repetitive structures, shafts and tanks with uniform walls, or indoor spaces dominated by weakly textured surfaces. Such inspection routes may necessarily cross extended regions with insufficient distinctive visual features.

We consider localization using only onboard cameras and an inertial measurement unit (IMU), without prior maps or external positioning infrastructure. Light detection and ranging (LiDAR) and other complementary sensors are outside this setup. Although multi-modal sensor fusion~\cite{Zhang2023, Wang2025, Yi2025} and deep learning-based feature extraction~\cite{Hu2025} have advanced localization, additional sensing changes the hardware configuration, while image processing remains limited by the available visual information. Without usable visual constraints, visual-inertial odometry (VIO) relies on inertial propagation and cannot guarantee reliable localization through prolonged visual deprivation. Perception-aware motion helps only when informative viewpoints remain accessible along the inspection route. %Consequently, even sophisticated visual-inertial estimators cannot guarantee reliable localization through prolonged visual deprivation.

These challenges are particularly relevant to underground inspection, where tunnels, mines, and damaged structures can combine limited visual distinctiveness with airborne dust affecting both cameras and LiDARs~\cite{GonzalezMarin2025, MontesGrova2026}. Geometric degeneracy can also challenge LiDAR-inertial odometry in tunnels, while dust and poor lighting impair VIO~\cite{GonzalezMarin2025}. Propeller downwash can further reduce visibility by resuspending dust\cite{MontesGrova2026}. Here, we specifically address insufficient visual texture under visibility and illumination conditions that still allow newly introduced features to be observed.

\begin{figure}[t]
    \centering
    \includegraphics[width=\linewidth]{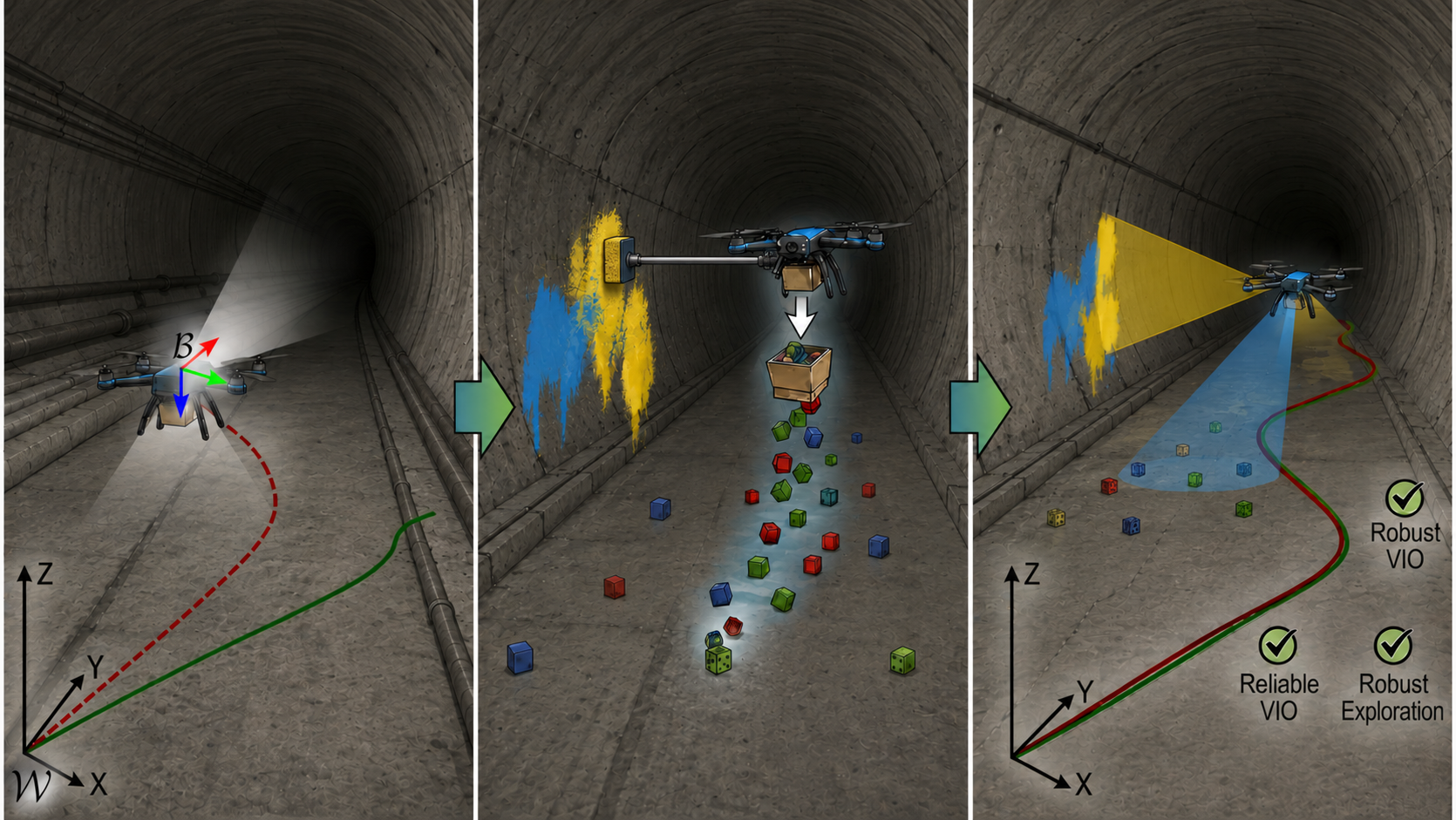}
    \caption{Conceptual illustration of VIO drift mitigation through on-demand visual feature creation. Left: repetitive structures and insufficient visual texture can degrade VIO and increase localization drift. Centre: the UAV introduces additional visual features by releasing objects onto the ground or marking accessible walls. Right: the added features provide new visual constraints, supporting recovery of VIO estimation quality and limiting further drift accumulation.}
    \label{fig:concept}
    \vspace{-0.3cm}
\end{figure}

Inspired by Tom Thumb's trail of pebbles, our UAV introduces persistent visual features by depositing ground markers or marking accessible walls through physical contact (Fig.~\ref{fig:concept}). These features provide additional visual constraints to support estimation and limit further drift accumulation.

We propose an integrated framework that couples online VIO quality monitoring with on-demand feature creation. Motivated by the information-based analysis of estimation degeneracy in~\cite{Zhang2016}, we construct an operational degradation indicator from the extended Kalman filter (EKF) information matrix and use temporal confirmation and hysteresis to trigger interventions. The robot selects a marking action according to the available surfaces and enabled mechanisms, and monitors the subsequent estimation response. 

Hence, our contribution is the integration of degradation monitoring, physical feature creation, and navigation within a camera-and-IMU localization framework for aerial inspection in visually degraded environments.

%The remainder of this paper is organized as follows. Section~\ref{sec2} reviews relevant literature. Section~\ref{sec3}  presents the proposed framework, while Section~\ref{sec4} discusses different case studies. Section~\ref{sec5}, finally, concludes the paper.
\section{Related Works}\label{sec2}

VIO has become the prevailing approach to localizing UAVs in GNSS-denied environments, whose effectiveness depends largely on the presence of visual features. For scenarios with abundant textures, the state-of-the-art VIO systems, including VINS-Mono~\cite{8421746}, provide good accuracy, whereas for scenarios with low textures, repetitive patterns, insufficient lighting, and homogeneity, feature detection will become increasingly difficult, leading to unbounded error and possible localization failure~\cite{Pan2024, Hu2025}. While LiDAR-based solutions have shown much promise in tunnel-like scenarios, the feasibility of adopting such solutions on light UAVs faces limitations of payload, cost, and power efficiency. Besides, as pointed out in~\cite{MontesGrova2026}, the parameter settings in LiDAR-Inertial Odometry are important, and its robustness against dust and geometric degeneracy must still be improved. The main limitation shared by these methods is the passive characteristic: they depend only on existing environmental features and cannot create features if there are none.

To overcome single-sensor limitations, the community has explored multi‑modal sensor fusion. Tightly coupled architectures that blend cameras, IMUs, and LiDARs at the raw data level offer robustness and accuracy, but at the cost of increased computational load and weight~\cite{Zhang2023, Wang2025}. Moreover, as noted in~\cite{Shi2025}, the real challenge is not only accuracy but \textit{uncertainty awareness} — knowing when sensors are unreliable. Yet even the best fusion cannot create information in a featureless void.

\textit{Active perception} plans motion to improve visual information, using deep reinforcement learning to avoid perceptually poor areas~\cite{Bartolomei2021} or perception-aware planning to favour texture-rich regions~\cite{Costante2017}. Its effectiveness, however, depends on the availability of informative viewpoints, which may be limited along feature-degraded inspection routes.

Visual place recognition supports loop closure, with learned descriptors improving robustness to illumination changes and occlusions~\cite{Sunderhauf2016, Masone2021}, although distinctive visual cues remain necessary~\cite{Tian2022}. Underground localization studies also highlight geometric degeneracy and the potential of thermal imaging under poor visibility~\cite{GonzalezMarin2025}, while intensity-based dust filtering is incorporated into autonomous exploration~\cite{MontesGrova2026}. These methods exploit existing observations, whereas our framework physically introduces additional visual features.

Artificial landmarks support reliable pose estimation~\cite{Kalaitzakis2021}, including QR codes as global anchors in LiDAR-inertial SLAM~\cite{Yang2026} and infrared beacons~\cite{LI2025115917}. Our work instead integrates \emph{active, on-demand feature creation} into exploration through ground marker deployment and wall marking.

For wall marking, we initially considered spray painting inspired by PaintCopter~\cite{Stilinovic2018}, but pressurized canisters, downwash-induced spray dispersion, and environmental concerns despite drift reducers~\cite{Steponavicius2025} motivated contact-based marking with a sponge. Contact is detected using an external force estimator~\cite{Ruggiero2015, 8059847} relying only on onboard IMU measurements and control inputs. Combined with an onboard marker dispenser, this enables physical feature creation without pre-installed infrastructure.
\begin{figure*}[t]
    \centering
\includegraphics[trim={0.5cm 1.5cm 0.0cm 1.0cm},clip,width=\linewidth]{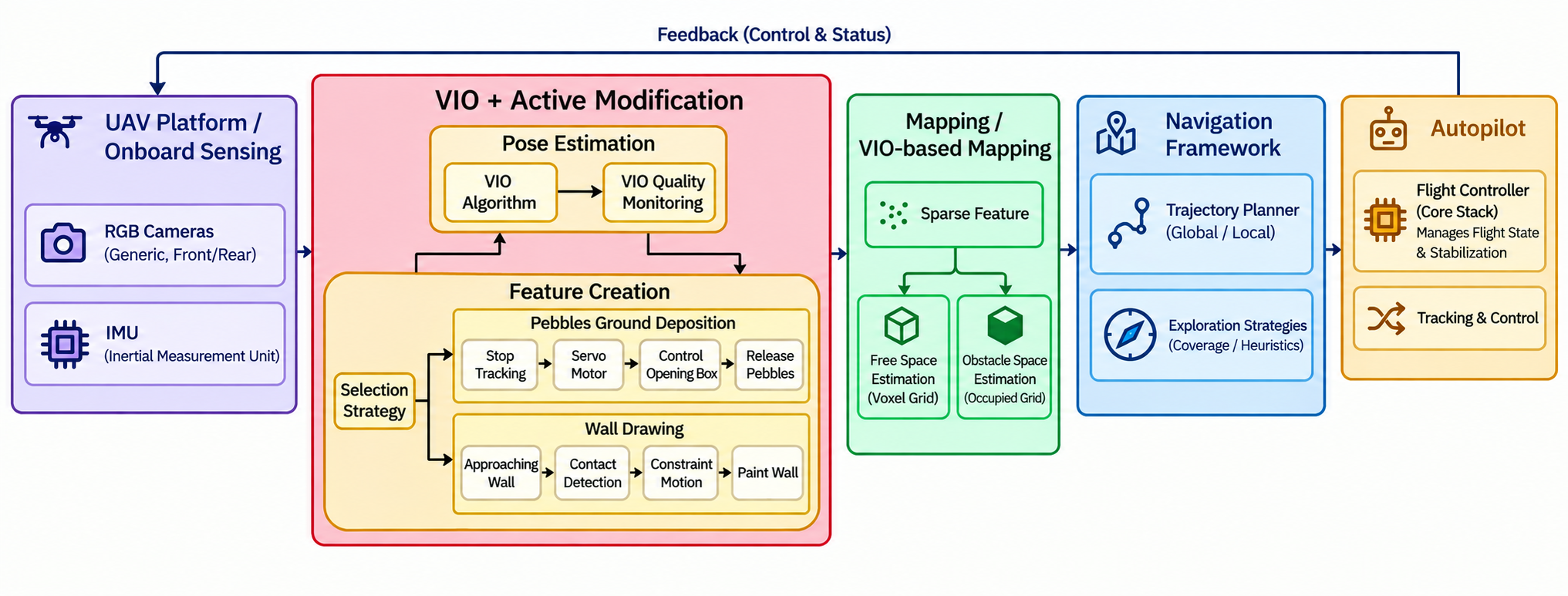}
\caption{Overview of the proposed VIO-based framework. Onboard sensors feed the VIO algorithm, whose quality is continuously monitored to trigger active marking through either ground pebble deposition or wall drawing. The VIO sparse visual information is then used to build a map, on which the navigation framework in~\cite{drones10040232} generates trajectories continuously tracked by the autopilot, closing the control loop with the sensing units.}  \label{fig:workflow}
\vspace{-0.3cm}
\end{figure*}
\section{Methodology}\label{sec3}
Consider a UAV whose task is to operate in an unknown environment. Only its maximum dimensions are known a priori; no prior map and no external positioning infrastructure are available. In such a setting, VIO serves as the primary state estimation source — yet it is precisely in these unknown, unstructured environments that visual features are most likely to be scarce or degraded. 

With reference to Fig.~\ref{fig:concept}, let $\mathcal{W}$ be the world inertial frame, fixed to the environment, and let $\mathcal{B}$ be the body frame attached to the robot's center of mass. The position ${p} \in \mathbb{R}^3$ and orientation ${R} \in SO(3)$ of $\mathcal{B}$ in $\mathcal{W}$ define the robot's pose, with ${v} \in \mathbb{R}^3$ and ${\omega} \in \mathbb{R}^3$ its linear and angular velocities. To perceive the environment, the robot is equipped with two camera sensors with different viewpoints: one oriented forward for navigation, and one tilted downward/backward to observe the ground facilitating marker detection. Let $\mathcal{C}_{\text{front}}$ and $\mathcal{C}_{\text{rear}}$ be the corresponding camera frames, rigidly attached to $\mathcal{B}$ with known extrinsics.

The problem statement of this work is to design a reliable solution to mitigate VIO drift based on active environment modification strategies that enable the UAV to deliberately generate visual features on demand during exploration. This paper focuses on two approaches: \textit{ground deposition}, releasing pebbles onto the floor as new visual landmarks, and \textit{wall drawing}, making physical contact with vertical surfaces to create persistent features. Both are integrated into a complete VIO-based exploration framework, detailed in this section.
\subsection{Workflow}\label{sec:overall_workflow}
We build upon an existing autonomous navigation framework~\cite{drones10040232}, which provides general mapping, planning, and control capabilities for UAVs in GNSS-denied environments. To this base, we add three key components: $(i)$ OpenVINS~\cite{9196524} as the baseline VIO estimator, $(ii)$ a lightweight mapping pipeline that converts sparse OpenVINS landmarks into an OctoMap for obstacle avoidance, and $(iii)$ a VIO quality monitoring pipeline triggering on-demand feature creation strategies.

The front and rear camera streams are processed by a single OpenVINS instance configured to handle both monocular cameras simultaneously. %The quality-monitoring logic is applied directly to the resulting odometry estimate. 
The same estimation and monitoring architecture is used in both simulation and experiments. Moreover, the complete system operates in a closed loop between perception and action. When the degradation metric indicates unreliable localization, the strategy selector determines which active modification behaviour to trigger. The choice between ground deposition and wall drawing is not pre-established; it depends on the target application context and accessibility of adjacent surfaces. In the case of tightly enclosed environments with many walls, such as tunnels, tanks or sewage pipes, wall drawing appears to be the favored approach, because vertical and curvilinear surfaces are always available and produce durable and contrasted marks. In contrast, for open or floor-oriented environments with few or inaccessible walls, ground deposition appears to be the natural option. In the case when surfaces of both kinds are available, the two approaches can be used together: the selector can alternate between, or combine, pebble release and wall drawing in order to create as many directional markers as possible. After feature creation, the monitor assesses whether VIO
quality has recovered sufficiently to resume exploration,
completing the feedback loop illustrated by Fig.~\ref{fig:workflow}. %The main framework components are then detailed in the remainder of this section.

%The remainder of this section details each component: the baseline VIO (Section~\ref{sec:openvins}), the mapping pipeline (Section~\ref{sec:vio_mapping}), the degradation metric (Section~\ref{sec:degradation_metric}), and the two active strategies (Section~\ref{sec:active_modification}).
\subsection{Preliminaries on Visual-Inertial Odometry: OpenVINS}
\label{sec:openvins}
OpenVINS~\cite{9196524} is a modular EKF‑based visual‑inertial odometry framework. The filter state is defined on the manifold \(\mathcal{M} = \mathbb{H} \times \mathbb{R}^{12}\) and includes the orientation quaternion \(\bar{q} \in \mathbb{H}\), position \(p\), velocity \(v\), and sensor biases \({b}_g, {b}_a \in \mathbb{R}^3\).
\begin{comment}
\begin{equation}
{x} = \left[ \bar{q}^\top \;\; {p}^\top \;\; {v}^\top \;\; {b}_g^\top \;\; {b}_a^\top \right]^\top.
\label{eq:openvins_state}
\end{equation}
\end{comment}
For brevity, we report the inertial component of the OpenVINS state only. The implemented filter additionally maintains a sliding window of cloned poses and the other variables enabled by its configuration. The filter propagates the state using IMU measurements \({a}_m,\omega_m\in\mathbb{R}^3\) (corrupted by biases and noise) according to the kinematics
\begin{align}
\dot{\bar{q}} &= \frac{1}{2} \bar{q} \otimes \begin{bmatrix} 0 \\ \omega_m - {b}_g \end{bmatrix}, \quad
\dot{{p}} = {v}, \quad\\
\dot{{v}} = {R}(\bar{q})(&{a}_m - {b}_a) + g{e}_3, \quad
\dot{{b}}_g = {0}, \quad
\dot{{b}}_a = {0},
\end{align}
where \(g\) is the gravity acceleration, \(e_3\in\mathbb{R}^3\) is the versor of the world-frame \(z\)-axis, and \(\otimes\) denotes the quaternion product (Hamilton product) between two quaternions.

Visual measurements are sparse feature tracks. For a feature observed in the camera frame \(\mathcal{C}\), the measurement model is
$
{z} = \pi\!\left( {R}_c \left( {R}^\top ({p}_f - {p}) - {t}_c \right) \right) + {n},
$
where \({p}_f \in \mathbb{R}^3\) is the landmark position in \(\mathcal{W}\), \({R} \in SO(3)\) is the rotation from \(\mathcal{B}\) to \(\mathcal{W}\), \({t}_c \in \mathbb{R}^3\) is the position of the camera origin expressed in \(\mathcal{B}\), \({R}_c \in SO(3)\) is the fixed rotation from \(\mathcal{B}\) to \(\mathcal{C}\), \(\pi(\cdot)\) is the camera projection function, and \({n} \sim \mathcal{N}(0, \sigma^2 {I})\). We disable loop closure and rely on the multi‑state constraint Kalman filter (MSCKF) for efficiency.

\subsection{VIO Mapping}
\label{sec:vio_mapping}

To support obstacle-aware exploration without relying on dense depth sensors, we complement the OpenVINS sparse landmarks with a lightweight mapping pipeline. Our implementation builds on the MonoSpheres approach~\cite{11517436}, re-engineered into a ROS~2/C++17 module for real-time performance and adapted to produce an OctoMap, a probabilistic occupancy map directly consumable by the planner in~\cite{drones10040232}.

New landmark observations are matched to existing ones via Mahalanobis gating and fused through a Bayesian update; each landmark retains its minimum observation distance as a reliability indicator. Landmarks are then projected into the image plane and triangulated via 2D Delaunay, yielding a sparse depth mesh. Connecting each mesh triangle to the camera center forms a free-space cone; the union of all such cones defines a visible free-space polyhedron. The resulting polyhedron approximates the visible free space inferred from the sparse landmark geometry. Its reliability depends on the validity of the interpolated surfaces and on the absence of unobserved obstacles within the reconstructed volume. To fill coverage gaps, an optional Open-Area Virtual Depth Estimation (OVDE) step injects synthetic points in regions with sufficient historical parallax and no known occlusions, while a Distance-Based Obstacle Filtering (DBOF) step prunes redundant landmarks, retaining the closest observation among nearby duplicates and permanently protecting any landmark ever observed near the camera. The free-space polyhedron is then sampled to mark free voxels, and the retained landmarks are inserted as occupied voxels, yielding the final OctoMap. As an extension to the original MonoSpheres pipeline, our implementation includes a wall-densifier module that detects planar surfaces in the sparse point cloud and generates additional synthetic points on them, increasing map density in structured environments such as corridors or tunnels.

\subsection{VIO Degradation Metric}\label{sec:degradation_metric}
To monitor the VIO estimate quality online, we adopt a degeneracy metric derived from the information matrix of the EKF-based estimation problem, following~\cite{Zhang2016}. The objective is to detect when the VIO estimation becomes unreliable due to texture scarcity, repetitive structures, or feature degradation, and trigger the active environment modification strategies described in Section~\ref{sec:active_modification}.

With reference to the EKF update formulation in Section~\ref{sec:openvins}, the original non-linear estimation problem is linearized at the current state estimate by computing the Jacobian of the measurement model, yielding the weighted linear least-squares formulation
\begin{equation}
\underset{\tilde{{x}}}{\arg\min} \left( \sum_{i=1}^{m} \left\| {y}_i - {H}_i \tilde{{x}} \right\|_{{\Sigma}_i}^2 + \left\| \tilde{{x}} \right\|_{{P}}^2 \right),
\label{eq:ekf_optimization}
\end{equation}
where $y_i = z_i - h_i(\hat{x}) \in \mathbb{R}^2$ is the reprojection residual of the $i$-th tracked feature, $m$ is the number of tracked features, ${H}_i\in\mathbb{R}^{2\times n}$ is the corresponding measurement Jacobian, and ${\Sigma}_i\in\mathbb{R}^{2\times 2}$ is the measurement noise covariance. The symbol $\tilde{x}\in\mathbb{R}^{n}$ denotes the local error-state perturbation, and we use the shorthand $\|r\|_S^2 = r^\top S^{-1} r$. The second term is the prior from the propagated state covariance ${P}\in\mathbb{R}^{n\times n}$. Following~\cite{Zhang2016}, the corresponding information matrix is
\begin{equation}
{J} = {H}^\top {\Sigma}^{-1} {H} + {P}^{-1},
\label{eq:information_matrix}
\end{equation}
where $H=[H_1^\top,\ldots,H_m^\top]^\top \in \mathbb{R}^{2m\times n}$ and $\Sigma=\operatorname{blkdiag}(\Sigma_1,\ldots,\Sigma_m) \in \mathbb{R}^{2m\times 2m}$, assuming independent measurement noise. We extract the position block ${J}_{pp} \in \mathbb{R}^{3 \times 3}$ and use its eigenvalues as an operational degradation indicator. This block describes conditional position information, with the remaining state variables fixed. When the VIO degrades (e.g., few tracked features, poor geometry, or motion degeneracy), information along one or more directions decreases, causing the corresponding eigenvalue to drop~\cite{Huang2010, Zhang2016}. The three eigenvalues, computed online and published at each estimator update, constitute the degradation metric used to drive the quality-monitoring logic.

A hysteresis-based algorithm, as in~\cite{PRITZL2023104315}, counters false positives from temporary variations, using a lower threshold $K_{j,1}$ and an upper threshold $K_{j,2}$ for each eigenvalue $\lambda_j$, $j\in \{1,2,3\}$, with $K_{j,1} < K_{j,2}$. 
%In addition, if all three eigenvalues remain nearly constant across consecutive updates (variation below a small tolerance $\varepsilon$) for more than $N_{\text{stag}}$ consecutive updates, the estimate is flagged as \emph{stagnant}, since no new information is being gathered despite the eigenvalues not having dropped below $K_{j,1}$.
A stagnation flag is raised when all three eigenvalues vary by less than $\varepsilon$ for more than $N_{\mathrm{stag}}$ consecutive updates. This provides an additional heuristic warning even when the eigenvalues remain above their lower thresholds.

The system evolves through four stages. From \texttt{NOM} (nominal), the
state escalates to \texttt{SUSP} (suspected degradation) as soon as
\emph{any} of the three eigenvalues $\lambda_j$ of ${J}_{pp}$ satisfies
$\lambda_j < K_{j,1}$,
or the stagnation condition is met. If all eigenvalues exceed their respective $K_{j,1}$ and
the stagnation condition clears before the confirmation
window $\Delta t$ elapses, the state reverts to \texttt{NOM};
otherwise it escalates to \texttt{DEGR}. 
Symmetrically, from \texttt{DEGR} the state
moves to \texttt{REC} (recovering) once \emph{all} eigenvalues satisfy
$\lambda_j > K_{j,2}$ (and the estimate is no longer stagnant),
confirming back to \texttt{NOM} if this condition persists for
$\Delta t$, or reverting to \texttt{DEGR} otherwise. The OR-logic on
degradation and AND-logic on recovery, combined with the
double-threshold, double-confirmation scheme, prevent chattering near
the degeneracy boundary in both directions. The resulting state is
published as the VIO quality status and serves as the input to the
strategy selector (see Fig.~\ref{fig:workflow}).

\subsection{Active Environment Modification Strategies}
\label{sec:active_modification}

The strategy selector is activated when the monitor enters \texttt{SUSP} in anticipatory mode or \texttt{DEGR} in reactive mode.
The selector evaluates the current environment -- that is, whether there are close walls, the ability to access the ground, and the direction associated with the degraded eigenvalue(s) -- and chooses between two complementary actions: \textit{ground deposition} (releasing pebbles on the floor) and \textit{wall drawing} (depositing visible marks on vertical walls). Both strategies are triggered online and executed autonomously, without requiring human intervention or pre-installed infrastructure.

\subsubsection{Ground Deposition}
\label{sec:ground_deposition}

The ground-deposition strategy consists of releasing small objects onto the floor below the UAV. When the strategy selector activates this behaviour, autonomous exploration is paused, a release command is issued, and a dispenser mounted underneath the drone opens and closes with a fixed time step. The released objects fall to the ground and act as artificial visual landmarks. After the release, a short confirmation window checks whether the new feature has been integrated; if VIO quality improves, exploration resumes, otherwise the strategy is triggered again.

\subsubsection{Physical interaction for wall painting}
\label{sec:force_estimation}

For wall painting, two rigid sticks ending with a sponge are attached to the UAV.
When wall drawing is selected, exploration is paused and velocity commands guide the UAV towards the selected wall. The approach requires the current state estimate and local map to remain sufficiently accurate for the manoeuvre. Anticipatory triggering initiates the action at the first degradation warning, whereas reactive triggering waits for temporal confirmation and may therefore initiate the approach with a more degraded estimate. An acceleration-based external force estimator~\cite{Ruggiero2015, 8059847} continuously monitors the estimated external force \(\hat{{f}}_{\text{ext}} \in \mathbb{R}^3\) using only the onboard IMU and control inputs, following the approach of~\cite{8059847}
$
\dot{\hat{{f}}}_{\text{ext}} = {K}_I \left( m {R} {a} - u_T {R} {e}_3 - \hat{{f}}_{\text{ext}} \right),
$
where \({K}_I \in \mathbb{R}^{3\times3}\) is a positive-definite diagonal gain matrix, \(m\) is the system mass, \({a} \in \mathbb{R}^3\) is the body-frame acceleration from the IMU (already accounting for gravity), and \(u_T\) is the total thrust. Contact is detected when the norm of the high-pass filtered external force exceeds a threshold \(f_c\) (see~\cite[Sec. VI]{8059847}). Upon contact confirmation, the velocity commands are interrupted, allowing the UAV to remain in contact with the wall and deposit a visible mark using the sponge. After a brief painting phase, the robot resumes exploration. The newly painted mark provides a fresh visual feature for the VIO, which is expected to recover reliable localization — closing the loop and allowing the mission to continue.
\begin{figure}[t]
    \centering
    \includegraphics[width=\linewidth]{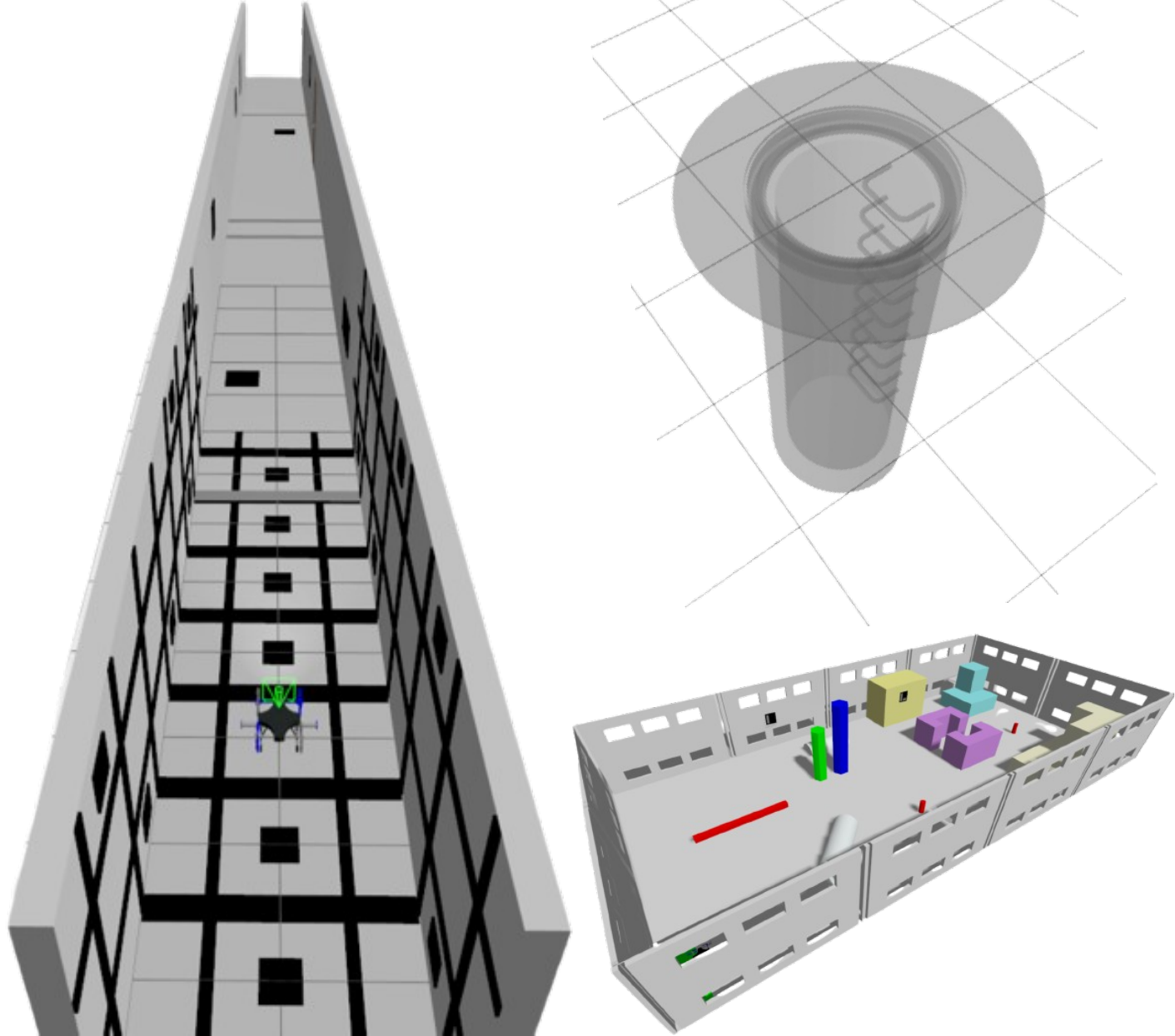}
\caption{Simulated scenarios: a long corridor/tunnel, a sewer shaft, and an open indoor space with sparse obstacles.}   \label{fig:sim}
\vspace{-0.3cm}
\end{figure}

\section{Results}\label{sec4}
The proposed framework has been validated both in simulation and on a real aerial platform and released online\footnote{Github repo:  \textbf{Code released upon paper acceptance.}} to support reproducibility and foster further research. The stack runs on ROS~2 Humble, with PX4 as the flight controller. At each step, the VIO mapping module (Section~\ref{sec:vio_mapping}) identifies free space from the OpenVINS landmarks. Within this free space, a random target pose is sampled. The navigation stack in~\cite{drones10040232} then generates a collision-free trajectory and commands the UAV to reach the target. During exploration, the degradation metric continuously monitors VIO quality. If degradation is detected, the exploration is paused, the active marking strategy is triggered, and once the VIO recovers, exploration resumes.

\subsection{Simulation Results and Analysis}
\label{sec:sim_results}

We conduct an ablation study (our method through on-demand feature creation (ODFC) versus VIO baseline) across three simulated environments (see Fig.~\ref{fig:sim}) selected to isolate the effectiveness of each strategy under specific conditions. In the open space, distant walls make ground deposition the natural choice; in the vertical sewer shaft, ground deposition is impractical, leaving wall drawing as the only viable strategy; in the corridor/tunnel, both strategies can be combined, providing a comprehensive test of the complete framework. In all scenarios, the degradation metric thresholds are fixed at $K_{j,1}=350$ and $K_{j,2}=750$, while cruise speed (0.5, 1.0, and 3.0 m/s) and triggering mode — \texttt{SUSP} (anticipatory, at the first degradation warning) versus \texttt{DEGR} (reactive, after temporal confirmation) — are varied. We evaluated performance using the position RMSE (root mean square of the position error over the full trajectory), the maximum position error, and the final position error, all computed after synchronizing estimated and ground-truth positions in a common reference frame for each successful trial. In the Gazebo simulator, coloured patches are spawned in the environment to emulate the appearance of the artificial features, enabling controlled evaluation of the active marking strategies without hardware dependencies.

\subsubsection{Sewer Scenario}
\label{sec:sewer}
The sewer environment consists of a vertical cylindrical shaft with an inner diameter of $1.30$~m and a depth of $6$~m, extending along the world $z$-axis. The shaft is featureless and texture-poor, with uniform curved walls that provide no distinctive visual landmarks. 
Ground deposition is unsuitable for supporting localization along the vertical route, whereas nearby walls allow contact-based feature creation.

Table~\ref{tab:sewer_results} compares baseline VIO and wall drawing at three cruise
speeds under anticipatory and reactive triggering. Wall drawing
reduced RMSE relative to the baseline in every tested condition.
At $0.5$~m/s, anticipatory triggering reduced RMSE from
$0.701$ to $0.397$~m, corresponding to approximately $43\%$.
The difference was larger at $3$~m/s, where baseline RMSE
reached $2.040$~m and the maximum error exceeded $10$~m,
while anticipatory wall drawing yielded $0.476$~m RMSE
and a maximum error of $0.569$~m.

Anticipatory triggering achieved lower RMSE than reactive
triggering at all three speeds, although both improved on
the baseline. These results support wall marking in confined
vertical environments and suggest that initiating the marking
manoeuvre before degradation is temporally confirmed can
improve estimation performance.
%The baseline drifts severely, especially at higher speeds: RMSE goes from $0.701$~m at $0.5$~m/s to $2.04$~m (max $10.953$~m) at $3$~m/s, confirming that passive VIO cannot sustain reliable localisation in such a texture-poor environment.
%Wall drawing reduced RMSE relative to the baseline at all tested speeds, with anticipatory triggering outperforming reactive triggering. At $v=3$~m/s, RMSE decreased from $2.040$~m to $0.476$~m with anticipatory triggering, compared with $0.530$~m with reactive triggering.
\begin{table}[t]
    \centering
    \caption{Sewer scenario position errors [m].}
    \label{tab:sewer_results}

    \begin{tabular}{ccccccc}
        \toprule
        \multicolumn{3}{c}{\textbf{Configuration}}
        & \textbf{Speed} & \textbf{RMSE}
        & \textbf{Max} & \textbf{Final} \\
        \midrule

        \multicolumn{3}{c}{\multirow{3}{*}{Baseline VIO}}
        & 0.5 m/s & 0.701 & 1.277 & 0.977 \\
        \multicolumn{3}{c}{}
        & 1.0 m/s & 0.550 & 1.780 & 1.780 \\
        \multicolumn{3}{c}{}
        & 3.0 m/s & 2.040 & 10.953 & 10.953 \\

        \midrule
        \multirow{6}{*}{\rotatebox[origin=c]{90}{ODFC}}
        & \multirow{6}{*}{
            \rotatebox[origin=c]{90}{\textit{Wall Drawing}}
        }
        & \multirow{3}{*}{Anticipatory}
        & \textbf{0.5 m/s} & \textbf{0.397}
        & \textbf{0.472} & \textbf{0.421} \\
        & & & 1.0 m/s & 0.450 & 0.521 & 0.483 \\
        & & & 3.0 m/s & 0.476 & 0.569 & 0.558 \\

        \cmidrule(lr){3-7}
        & & \multirow{3}{*}{Reactive}
        & 0.5 m/s & 0.437 & 0.626 & 0.444 \\
        & & & 1.0 m/s & 0.499 & 0.563 & 0.560 \\
        & & & 3.0 m/s & 0.530 & 0.597 & 0.573 \\

        \bottomrule
    \end{tabular}
    \vspace{-0.3cm}
\end{table}

\subsubsection{Indoor Open Space Scenario}
\label{sec:warehouse}
The environment comprises a $20\times10$~m indoor area with
sparse obstacles and moderate texture. Although natural
features are present, their distribution and distance from
the UAV can limit useful visual constraints. Reaching distant
walls would require additional travel before feature creation;
ground deposition is therefore selected as the active strategy.
\begin{table}[b]
    \centering
    \caption{Open space scenario position errors [m].}
    \label{tab:open_space_results}

    \begin{tabular}{ccccccc}
        \toprule
        \multicolumn{3}{c}{\textbf{Configuration}}
        & \textbf{Speed} & \textbf{RMSE}
        & \textbf{Max} & \textbf{Final} \\
        \midrule

        \multicolumn{3}{c}{\multirow{3}{*}{Baseline VIO}}
        & 0.5 m/s & 0.890 & 4.623 & 4.623 \\
        \multicolumn{3}{c}{}
        & 1.0 m/s & 0.758 & 4.552 & 4.552 \\
        \multicolumn{3}{c}{}
        & 3.0 m/s & 17.719 & 136.338 & 136.338 \\

        \midrule
        \multirow{6}{*}{\rotatebox[origin=c]{90}{ODFC}}
        & \multirow{6}{*}{
            \rotatebox[origin=c]{90}{\textit{Ground Depos.}}
        }
        & \multirow{3}{*}{Anticipatory}
        & \textbf{0.5 m/s} & \textbf{0.106}
        & \textbf{0.218} & \textbf{0.135} \\
        & & & 1.0 m/s & 0.208 & 0.418 & 0.404 \\
        & & & 3.0 m/s & 0.173 & 0.385 & 0.095 \\

        \cmidrule(lr){3-7}
        & & \multirow{3}{*}{Reactive}
        & 0.5 m/s & 0.234 & 0.370 & 0.369 \\
        & & & 1.0 m/s & 0.125 & 0.249 & 0.124 \\
        & & & 3.0 m/s & 0.174 & 0.373 & 0.209 \\

        \bottomrule
    \end{tabular}
    \vspace{-0.3cm}
\end{table}
%Table~\ref{tab:open_space_results} compares the position errors (RMSE, max, final) for the two methodologies at three cruise speeds, with both triggering modes. The simulation reveals that even when the environment provides sparse natural features, passive VIO remains unable to maintain reliable localisation. At $0.5$~m/s, the error grows to $0.890$~m RMSE and exceeds $4.5$~m at the end of the run; at $1.0$~m/s, the final error is similarly large. At $3.0$~m/s, the estimate diverges completely, confirming that sparse natural texture alone — however present — cannot prevent unbounded drift in open spaces without some form of active intervention.
Table~\ref{tab:open_space_results} reports the position errors for baseline VIO and
ground deposition under both triggering modes. Baseline RMSE
was $0.890$ and $0.758$~m at $0.5$ and $1$~m/s,
respectively, with final errors exceeding $4.5$~m in both
cases. At $3$~m/s, RMSE increased to $17.719$~m,
indicating severe estimation degradation in the reported trial.
Ground deposition reduced all three reported error metrics
relative to the baseline at every speed. Anticipatory
triggering yielded $0.106$~m RMSE at $0.5$~m/s,
approximately $88\%$ below the baseline. However, its
advantage over reactive triggering was not uniform:
at $1$~m/s, reactive triggering achieved lower RMSE
($0.125$ versus $0.208$~m), whereas at $3$~m/s
the two modes produced similar values ($0.174$ and
$0.173$~m, respectively). Thus, ground deposition was
effective throughout the tested speed range, while the
relative benefit of anticipatory triggering depended on
the operating condition.

%Ground deposition drastically improves performance across all speeds. The anticipatory mode achieves the best results: at 0.5 m/s, the RMSE is $0.106$~m (max $0.218$~m, final $0.135$~m), representing an $88\%$ reduction compared to the baseline. Even at $3.0$~m/s, the anticipatory mode maintains bounded drift ($0.173$~m RMSE, max $0.385$~m, final $0.095$~m), demonstrating that ground deposition is highly effective in open spaces and that the conservative triggering mode performs reliably even under aggressive motion.

%The reactive mode performs better than the baseline, with RMSE values of $0.234$~m at $0.5$~m/s and $0.174$~m at $3.0$~m/s. Notably, the mode avoids failure in this environment, having manoeuvring room for interventions. Overall, ground deposition reduced the reported position errors relative to the baseline at all tested speeds. Anticipatory triggering yielded lower RMSE at $0.5$ and $3.0$~m/s, whereas reactive triggering achieved lower errors at $1.0$~m/s.

\subsubsection{Corridor/Tunnel Scenario}
\label{sec:corridor}

The last scenario features a long ($33\times5$~m), feature-degraded corridor in which both the floor and nearby walls are
available for feature creation. This scenario allows the
individual marking strategies to be compared with their
combined use. 

Table~\ref{tab:corridor_results} reports the corresponding position errors, including anticipatory and reactive triggering
for the complete framework.
In the reported trials, baseline RMSE increased from
$0.557$~m at $0.5$~m/s to $0.878$~m at $3$~m/s.
Both individual strategies reduced RMSE at every speed,
but neither consistently outperformed the other. Wall
drawing achieved lower RMSE at $0.5$ and $1$~m/s,
whereas ground deposition performed better at $3$~m/s.
The combined anticipatory configuration yielded the lowest
RMSE at each speed, reaching $0.064$, $0.106$, and
$0.218$~m at $0.5$, $1$, and $3$~m/s, respectively.
At the highest speed, its maximum error was $0.417$~m,
compared with $11.199$~m for the combined reactive
configuration. This difference supports investigating early
intervention, particularly when delayed marking must be
executed using an already degraded estimate.

These comparisons describe the trials reported in
Table~\ref{tab:corridor_results} and do not establish uniform superiority across
randomized missions. The following factorial campaign
therefore evaluates the effects of velocity and strategy
activation over repeated trials, distinguishing divergence
probability from accuracy among non-divergent runs.

%The complete framework with the anticipatory mode performs best across all tested speeds. At 0.5~m/s it achieves an RMSE of 0.064~m (max 0.146~m), an 88\% reduction compared to the baseline. Even at 3.0~m/s, where things get harder, the anticipatory mode keeps drift bounded (0.218~m RMSE, max 0.417~m), showing that intervening early allows the system to recover gracefully. The reactive mode, by contrast, performs reasonably well at low speeds ($0.288$~m RMSE at $0.5$~m/s, $0.177$~m at $1.0$~m/s), but fails at $3.0$~m/s, reaching a maximum error of $11.199$~m. The reactive state is reached only after the estimate has significantly degraded; intervention at this stage is not only ineffective but often destabilising, as poorly-timed feature creation attempts compound the error, ultimately causing instability and crashes. This highlights the critical role of early intervention in preventing irreversible drift.

\begin{table}[t]
    \centering
    \caption{Corridor scenario position errors [m].}
    \label{tab:corridor_results}

    \begin{tabular}{ccccccc}
        \toprule
        \multicolumn{3}{c}{\textbf{Configuration}}
        & \textbf{Speed} & \textbf{RMSE}
        & \textbf{Max} & \textbf{Final} \\
        \midrule

        \multicolumn{3}{c}{\multirow{3}{*}{Baseline VIO}}
        & 0.5 m/s & 0.557 & 1.266 & 0.766 \\
        \multicolumn{3}{c}{}
        & 1.0 m/s & 0.620 & 1.365 & 0.926 \\
        \multicolumn{3}{c}{}
        & 3.0 m/s & 0.878 & 1.965 & 1.434 \\

        \midrule
        \multicolumn{3}{c}{\multirow{3}{*}{ODFC -- Ground Deposition}}
        & 0.5 m/s & 0.208 & 0.484 & 0.254 \\
        \multicolumn{3}{c}{}
        & 1.0 m/s & 0.236 & 0.512 & 0.159 \\
        \multicolumn{3}{c}{}
        & 3.0 m/s & 0.398 & 1.303 & 0.848 \\

        \midrule
        \multicolumn{3}{c}{\multirow{3}{*}{ODFC -- Wall Drawing}}
        & 0.5 m/s & 0.184 & 0.442 & 0.191 \\
        \multicolumn{3}{c}{}
        & 1.0 m/s & 0.158 & 0.326 & 0.231 \\
        \multicolumn{3}{c}{}
        & 3.0 m/s & 0.463 & 2.498 & 2.296 \\

        \midrule
        \multirow{6}{*}{\rotatebox[origin=c]{90}{ODFC}}
        & \multirow{6}{*}{
            \rotatebox[origin=c]{90}{\textit{Whole Framework}}
        }
        & \multirow{3}{*}{Anticipatory}
        & \textbf{0.5 m/s} & \textbf{0.064}
        & \textbf{0.146} & \textbf{0.089} \\
        & & & 1.0 m/s & 0.106 & 0.209 & 0.146 \\
        & & & 3.0 m/s & 0.218 & 0.417 & 0.384 \\

        \cmidrule(lr){3-7}
        & & \multirow{3}{*}{Reactive}
        & 0.5 m/s & 0.288 & 0.723 & 0.511 \\
        & & & 1.0 m/s & 0.177 & 0.335 & 0.284 \\
        & & & 3.0 m/s & 1.494 & 11.199 & 11.199 \\

        \bottomrule
    \end{tabular}
    \vspace{-0.3cm}
\end{table}

%On average, the anticipatory mode achieves an RMSE of $0.129$~m across all speeds, compared to $0.653$~m for reactive mode and $0.685$~m for the baseline. These results confirm that early intervention, triggered as soon as the quality monitor enter \texttt{SUSP}, is essential for reliable VIO stabilisation across a wide range of speeds.

\subsection{Statistical Analysis of Simulation Results}
We conducted a \(3\times2\times2\) factorial simulation campaign in the corridor scenario, varying cruise speeds ($v$ equal to the three values already considered above), the activation of wall drawing ($W=\{1,0\}$, two levels, yes or no), and ground deposition ($G=\{1,0\}$, two levels, yes or no). Notice that $(W, G)=(0,0)$ corresponds to the Baseline VIO. We used the anticipatory trigger throughout, with $75$ trials per condition and 900 runs overall. In each trial, the UAV explored the corridor by repeatedly sampling goals from the reconstructed free space described in Section~\ref{sec:vio_mapping}. A trial was considered successful when the exploration reached the goal at the end of the corridor within the $600$~s time limit; otherwise, it terminated upon timeout. Randomness arose from target sampling and Gaussian sensor noise. Each simulation run used a seed initialized from the machine time, without deliberately reusing seeds across configurations.

Position RMSE was computed over each recorded trajectory. For statistical analysis, runs with RMSE above $1.5$~m were classified as divergent using an empirical post-hoc threshold. We analysed the probability of exceeding this threshold and, separately, the RMSE of the remaining runs. This operational measure of estimation robustness is distinct from mission completion within the $600$~s limit.

\paragraph{Robustness analysis.}
The trials comprised $305$ divergent and $595$ convergent runs, corresponding to an overall divergence rate of $33.89\%$. The observed divergence rate ranged from $4\%$ for $(v,W,G)=(0.5,1,1)$ to $76\%$ for $(v,W,G)=(3,0,0)$. A full-factorial logistic regression was fitted with divergence as a binary response and velocity, wall drawing, and ground deposition as categorical predictors, including all interaction terms. Each model term was assessed through a likelihood-ratio test against the corresponding reduced nested model. All three main effects were significant: velocity $v$ ($\chi^2_{2}=33.05$, $p<0.001$), wall drawing $W$ ($\chi^2_{1}=7.31$, $p=0.0068$), and ground deposition $G$ ($\chi^2_{1}=10.05$, $p=0.0015$), whereas none of the two-way or three-way interactions were significant ($p>0.13$). Post-hoc pairwise comparisons were performed on the marginal levels of each factor
(pooled over the other two), with Holm correction.

For velocity, the odds of divergence were significantly higher at $v=1$~m/s
than at $v=0.5$~m/s (odds ratio (OR) $=3.80$, 95\% confidence interval (CI): $2.59$--$5.59$,
$p_{\mathrm{Holm}} = 9.03\times10^{-12}$) and significantly higher at
$v=3$~m/s than at $v=0.5$~m/s (OR $=4.07$, 95\% CI: $2.77$--$5.99$,
$p_{\mathrm{Holm}} = 6.51\times10^{-13}$), whereas $v=1$ and $v=3$~m/s did
not differ significantly from each other (OR $=0.93$, 95\% CI:
$0.68$--$1.29$, $p_{\mathrm{Holm}}=0.680$). Wall drawing significantly
reduced the marginal odds of divergence relative to its absence
(OR $=0.37$, 95\% CI: $0.28$--$0.50$, $p_{\mathrm{Holm}} = 2.53\times10^{-11}$),
as did ground deposition (OR $=0.46$, 95\% CI: $0.35$--$0.61$,
$p_{\mathrm{Holm}} = 1.18\times10^{-7}$).

Marginal comparisons showed higher divergence odds at $v=1$ and
$v=3$~m/s than at $v=0.5$~m/s, with no significant difference between
$v=1$ and $v=3$~m/s. Enabling either marking strategy was associated
with lower marginal odds of divergence.
%Relative to the $v=0.5$~m/s reference condition, the odds of divergence increased by a factor of $4.52$ at $v=1$~m/s ($95\%$ Confidence Interval (CI): $2.28$--$8.98$) and by a factor of $6.33$ at $v=3$~m/s ($95\%$ CI: $3.10$--$12.95$). Conversely, wall drawing reduced the odds of divergence to $0.344$ times those observed without wall drawing (95\% CI: $0.154$--$0.765$), while ground deposition reduced them to $0.273$ times the corresponding odds ($95\%$ CI: $0.117$--$0.635$).

%Since no interaction was detected in the logistic model, pairwise comparisons were performed on the marginal levels of the significant factors. After Holm correction, both $v=1$ and $v=3$~m/s exhibited significantly higher divergence rates than $v=0.5$~m/s ($p_{\mathrm{Holm}}<0.001$ for both comparisons), whereas 1 versus 3~m/s was not significant ($p_{\mathrm{Holm}}=0.680$). Wall drawing and ground deposition both significantly reduced divergence ($p_{\mathrm{Holm}}<0.001$ for both comparisons). Overall, robustness was therefore strongly affected by commanded velocity, while the presence of wall and ground features was associated with lower divergence probability.

\paragraph{Conditional accuracy analysis.}
Conditional accuracy was evaluated on the $595$ convergent trials only. Because the RMSE distributions were not assumed to be Gaussian and the numbers of convergent trials were unbalanced across conditions, the Aligned Rank Transform (ART) was used to assess the factorial effects. No significant main effects were found for velocity ($F_{2,583}=0.22$, $p=0.800$, $\eta_p^2=0.0008$), wall drawing ($F_{1,583}=2.23$, $p=0.136$, $\eta_p^2=0.0038$), or ground deposition ($F_{1,583}=0.15$, $p=0.699$, $\eta_p^2=0.0003$). In contrast, both the velocity$\times$wall interaction ($F_{2,583}=10.08$, $p<0.001$, $\eta_p^2=0.033$) and the velocity$\times$wall$\times$ground interaction ($F_{2,583}=9.40$, $p<0.001$, $\eta_p^2=0.031$) were significant. The significant three-way interaction indicates that the effect of velocity on conditional tracking accuracy depends on both wall and ground configuration; therefore, the velocity$\times$wall interaction should not be interpreted independently of the ground condition.

The significant interactions were decomposed through simple-effect comparisons, conditioning each comparison on the remaining factors. Velocity levels were compared within each wall--ground configuration, while wall and ground levels were compared within each velocity--environment configuration. Pairwise comparisons used the Wilcoxon rank-sum test on the original convergent RMSE values, with Holm correction for multiple comparisons; the rank-biserial correlation ($r_{\mathrm{rb}}$) was used as the effect-size measure.

For velocity, significant differences were found at $(W,G)=(0,1)$ between $v=0.5$ and $v=1$~m/s ($p_{\mathrm{Holm}}=0.0051$, $r_{\mathrm{rb}}=0.405$) and between $v=1$ and $v=3$~m/s ($p_{\mathrm{Holm}}<0.001$, $r_{\mathrm{rb}}=-0.493$), at $(W,G)=(1,0)$ between $v=0.5$ and $v=1$~m/s ($p_{\mathrm{Holm}}<0.001$, $r_{\mathrm{rb}}=0.550$) and between $v=1$ and $v=3$~m/s ($p_{\mathrm{Holm}}=0.0329$, $r_{\mathrm{rb}}=-0.354$), and at $(W,G)=(1,1)$ between $v=1$ and $v=3$~m/s ($p_{\mathrm{Holm}}=0.0401$, $r_{\mathrm{rb}}=0.342$). No velocity comparison was significant for $(W,G)=(0,0)$. For wall drawing, a significant simple effect was found only at $v=1$~m/s and $G=1$, where the median RMSE increased from $0.127$ to $0.303$~m ($p_{\mathrm{Holm}}<0.001$, $r_{\mathrm{rb}}=-0.594$). Significant ground simple effects were observed at $(v,W)=(0.5,1)$ ($p_{\mathrm{Holm}}=0.0101$, $r_{\mathrm{rb}}=0.344$), $(v,W)=(1,0)$ ($p_{\mathrm{Holm}}=0.0092$, $r_{\mathrm{rb}}=0.517$), and $(v,W)=(1,1)$ ($p_{\mathrm{Holm}}<0.001$, $r_{\mathrm{rb}}=-0.475$).

\paragraph{Discussion}
The results distinguish two aspects of estimation performance:
the probability of exceeding the divergence threshold and
the accuracy of runs that remain below it. Lower cruise speed
and active marking were associated with reduced divergence,
with the lowest observed rate, $4\%$, occurring at
$(v,W,G)=(0.5,1,1)$. This supports combining the marking
mechanisms at low speed and avoiding large estimation
errors in the tested corridor.

Conditional accuracy showed a different pattern. At
$v=1$~m/s, enabling both strategies yielded higher median
RMSE than enabling either alone; with both enabled,
$v=3$~m/s yielded lower conditional RMSE than $v=1$~m/s.
Thus, neither lower speed nor additional marking mechanisms
uniformly improved accuracy among non-divergent runs.

These comparisons concern different subsets of retained
trials: a configuration with more divergent runs can still
show low RMSE among those that remain. Conditional RMSE
alone therefore cannot rank overall performance or establish
an inherent robustness--accuracy trade-off. The practical
implication is to assess divergence probability first and
conditional accuracy alongside it, rather than selecting a
configuration from RMSE alone. These findings apply to the
tested corridor and the adopted $1.5$~m threshold; the
factorial campaign evaluates anticipatory triggering and
does not establish its statistical superiority over reactive
triggering.

\subsection{Experimental Validation}

%Real-world experiments were conducted in a mock-up feature-degraded indoor corridor 
%(see Fig.~\ref{fig:setup_ex})
%, mimicking the setup in Section~\ref{sec:corridor}, which best represents the targeted feature-degraded conditions. The corridor walls and floor are entirely covered with white cloths, leaving no exploitable natural features. Differently from the closed-loop simulations, here we adopt an open-loop validation strategy. This choice follows a well-established practice in the evaluation of odometry algorithms, where the estimator under test runs in parallel to the control loop rather than being used as its feedback source~\cite{10058169}. This setup isolates estimation performance from the control loop, ensuring that any degradation in the odometry estimate is not masked by the controller's corrective actions. Accordingly, the exploration is driven by the OptiTrack motion capture system, which provides feedback to the flight controller, while OpenVINS runs in the background on the same data stream. The role of VIO is therefore not to control the UAV, but to act as a selector for the proposed active modification strategies: when the degradation metric detects that the VIO estimate is becoming inconsistent, the corresponding environment modification action is triggered. Whereas the VIO becomes inconsistent, the active strategy mitigates the problem, restoring consistency and improving the estimate by a measurable percentage, independently of the control loop.

\begin{figure}[t]
    \centering
 \includegraphics[width=\linewidth]{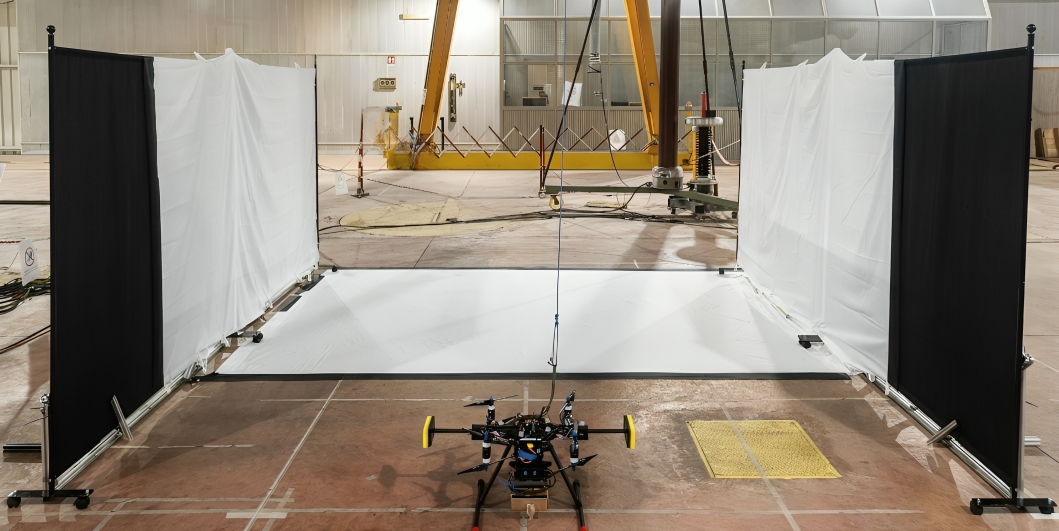}\\
\vspace{0.2cm}
    \includegraphics[trim={0.0cm 3cm 0.0cm 4.0cm},clip,width=\linewidth]{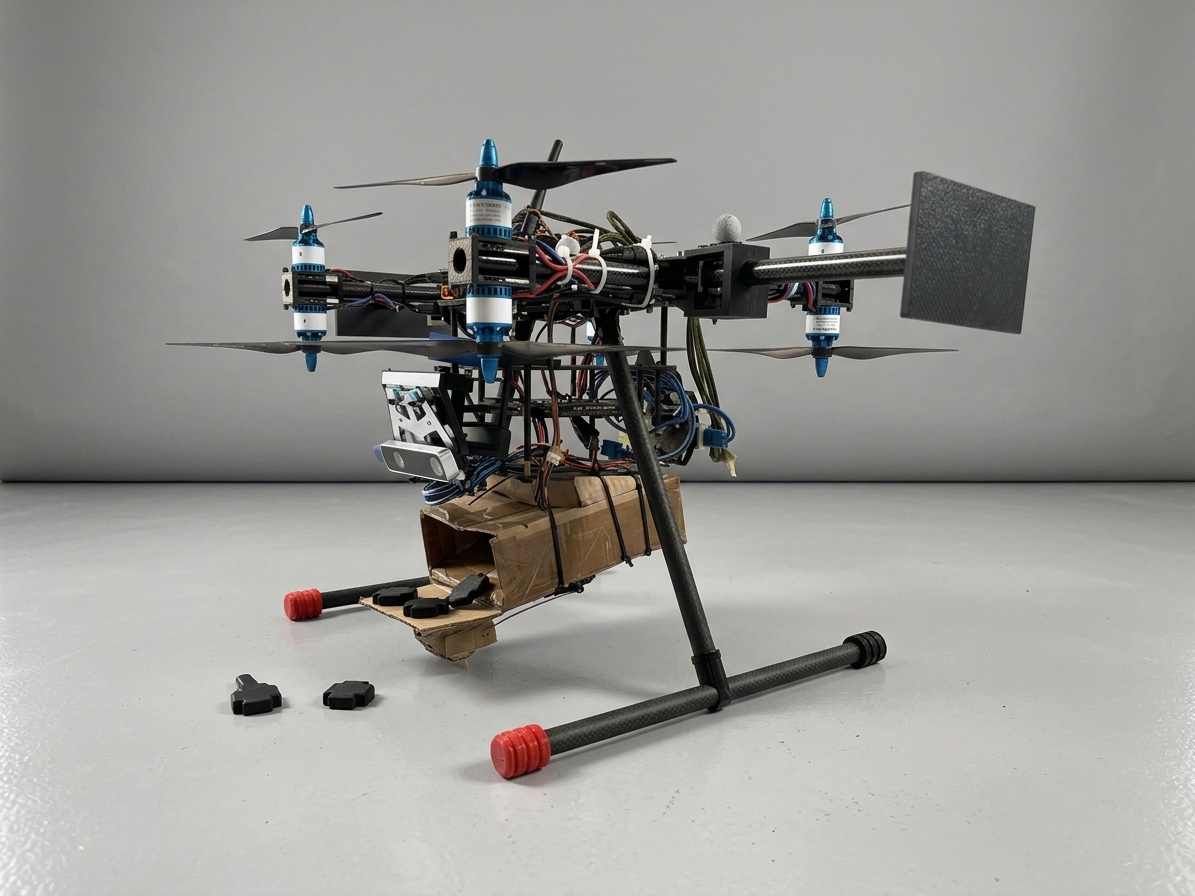}
    \caption{(Top) Experimental setup replicating the corridor scenario in Sec.~\ref{sec:corridor}. (Bottom) The UAV is equipped with two sponge end-effectors for wall drawing and a bottom-mounted, PWM-controlled box for releasing pebbles on demand.}\label{fig:setup_ex}
    \vspace{-0.3cm}
\end{figure}

Real-world experiments were conducted in a mock-up feature-degraded indoor corridor (see Fig.~\ref{fig:setup_ex}), mimicking the setup described in Sec.~\ref{sec:corridor}. The corridor walls and floor were fully covered with white cloth, providing no exploitable natural visual features. Given the results in Sec.~\ref{sec:sim_results}, we performed an ablation study comparing VIO Baseline versus our ODFC whole framework with Anticipatory triggering at a cruise speed of $0.5$~m/s.
Unlike the simulations, the experiments use OptiTrack feedback for flight control, while OpenVINS and the feature-creation logic run online. This choice follows a well-established practice in the evaluation of odometry algorithms, where the estimator under test runs in parallel to the control loop rather than being used as its feedback source~\cite{10058169}. This setup evaluates VIO under externally stabilized flight, while retaining the online perception--action loop that triggers feature creation.

The platform is a coaxial octacopter equipped with a Pixhawk~6C flight controller and two Intel RealSense T265 cameras (see Fig.~\ref{fig:setup_ex}). All processing was performed onboard on a LattePanda~3 Delta. Five runs were conducted for each configuration. Reported metrics are averaged across the five runs. For the baseline, the VIO accumulated an RMSE of $1.52$~m, a maximum error of $2.79$~m, and a final error of $2.5$~m over the corridor. With the complete framework, the RMSE decreased to $0.99$~m, with a maximum error of $1.79$~m and a final error of $1.24$~m. The average estimation error before feature deployment was $1.07$~m and dropped to $0.56$~m after deployment, corresponding to an average improvement of $47.6$\%. %Although smaller than in simulation the improvement confirms that the strategy reduces VIO error in a genuinely feature-degraded environment. The same software stack is used in both cases, with the discrepancy attributed to the sim-to-real gap and non-fine-tuned VIO; more extensive validation is left to future work.
These results indicate reduced VIO error in the tested corridor under externally stabilized flight. The magnitude of the improvement differed from simulation, although the same software architecture was used. The present experiments do not isolate the sources of this discrepancy although we believe it lays in the sim-to-real gap and/or non-fine-tuned VIO.

\begin{comment}
\begin{table}[t]
    \centering
    \caption{Experimental validation Comparison.}
    \begin{tabular}{
        >{\raggedright\arraybackslash}p{0.26\linewidth}
        >{\centering\arraybackslash}p{0.13\linewidth}
        >{\centering\arraybackslash}p{0.15\linewidth}
        >{\centering\arraybackslash}p{0.19\linewidth}
    }
        \toprule
        \textbf{Configuration} & \textbf{Speed} & \textbf{RMSE [m]} & \textbf{Improvement} \\
        \midrule
        Baseline VIO & 0.5 m/s & 1.6796 & -- \\
        \midrule
        Ground Deposition & 0.5 m/s & \textcolor{red}{TBD} & \textcolor{red}{TBD}\% \\
        Wall Drawing & 0.5 m/s & \textcolor{red}{TBD} & \textcolor{red}{TBD}\% \\
        \midrule
        Whole Framework & 0.5 m/s & \textcolor{red}{TBD} & \textcolor{red}{TBD}\% \\
        \bottomrule
    \end{tabular}
    \label{tab:exp_results}
\end{table}
\end{comment}

\section{Conclusions}\label{sec5}
This paper presented an active environment-modification
framework that couples online VIO quality monitoring with
ground marker deployment and contact-based wall marking.
The approach supplements insufficient natural texture with
visual cues created by the UAV, using cameras and an IMU
for localization.
Three simulated environments illustrated the use of the
two mechanisms under different surface-accessibility
conditions. A $900$-run factorial campaign associated lower
cruise speed and active marking with reduced divergence,
while revealing configuration-dependent accuracy among
non-divergent runs. These findings support evaluating both
outcomes when selecting a marking configuration.
Corridor experiments showed an approximately $35\%$
reduction in mean per-run position RMSE under OptiTrack-based
flight control, with VIO-driven marking decisions active
online. 

Future work will evaluate the complete framework
with VIO feedback for flight control and investigate
strategy selection that accounts for estimation quality,
surface accessibility, and the time and material costs
of each intervention.

%{\sout{Indeed, \textit{once upon a time}, it was believed that an autonomous agent could sustain long-term exploration based on VIO simply by observing the natural features of the environment. In practice, however, when those features deteriorated or disappeared, the agent was no longer able to locate itself in the absence of visual cues; with the proposed framework, on the other hand, it is able to localise itself in any situation, managing to live \textit{happily ever after}.}}

\section*{Acknowledgment}
ChatGPT (GPT-6 Astra) generated Figs.~\ref{fig:concept} and~\ref{fig:workflow}; Fig.~\ref{fig:setup_ex} is instead a polished version of the authors' real pictures.

\bibliographystyle{IEEEtran}
%\bibliography{bibliography}
\bibliography{Bibliography.bib}

\end{document}